\documentclass{article}
\usepackage{spconf,amsmath,graphicx}
\usepackage{amssymb}
\usepackage{mwe} 
\usepackage{booktabs}
\usepackage{amsfonts,bm}
\usepackage{multirow}
\usepackage{multicol}
\usepackage[normalem]{ulem} 
\newcommand{\ulgood}[1]{\uline{#1}}
\usepackage[normalem]{ulem} 

\title{U-Harmony: Enhancing Joint Training for Segmentation Models with Universal Harmonization}
\name{\parbox{\textwidth}{\centering 
    Weiwei Ma$^{\star}$  Xiaobing Yu$^{\star}$  Peijie Qiu$^{\star}$  Jin Yang$^{\star}$  Pan Xiao$^{\star}$  Xiaoqi Zhao$^{\dagger}$
    Xiaofeng Liu$^{\dagger}$  Tomo Miyazaki$^{\S}$  Shinichiro Omachi$^{\S}$  Yongsong Huang$^{\S}$
}}

\address{\vspace{-3mm}$^{\star}$ Washington University in St. Louis, 
    $^{\dagger}$ Yale University, 
    $^{\S}$ Tohoku University}

\begin{document}

\maketitle

\begin{abstract}
    In clinical practice, medical segmentation datasets are often limited and heterogeneous, with variations in modalities, protocols, and anatomical targets across institutions. Existing deep learning models struggle to jointly learn from such diverse data, often sacrificing either generalization or domain-specific knowledge. To overcome these challenges, we propose a joint training method called Universal Harmonization (U-Harmony), which can be integrated into deep learning-based architectures with a domain-gated head, enabling a single segmentation model to learn from heterogeneous datasets simultaneously. By integrating U-Harmony, our approach sequentially normalizes and then denormalizes feature distributions to mitigate domain-specific variations while preserving original dataset-specific knowledge. More appealingly, our framework also supports universal modality adaptation, allowing the seamless learning of new imaging modalities and anatomical classes. Extensive experiments on cross-institutional brain lesion datasets demonstrate the effectiveness of our approach, establishing a new benchmark for robust and adaptable 3D medical image segmentation models in real-world clinical settings.
\end{abstract}


\section{Introduction}

Despite the significant advances in 3D medical lesion segmentation driven by deep learning~\cite{litjens2017survey}, clinical translation remains hindered by a fundamental challenge known as domain shift. Models trained on a single domain exhibit poor generalization when confronted with the myriad variations in acquisition modalities, imaging protocols, and annotation standards endemic to real-world clinical data~\cite{kamnitsas2017unsupervised}. This heterogeneity is not a trivial exception but the norm. For instance, prominent datasets like UCSF-BMSR~\cite{ucsf_bmsr} and BraTS-METS 2023~\cite{brats_mets} possess irreconcilable differences in their included modalities and specific annotation goals.

To overcome this fragmentation, \textbf{Joint training} (JT) has been widely adopted as a paradigm for learning shared representations from diverse datasets, with the goal of enhancing generalization. Nevertheless, conventional JT approaches are brittle; they often fail to reconcile significant dataset heterogeneity. This paradoxically re-introduces the very performance-degrading domain shifts they were designed to eliminate.
 
While domain generalization (DG) often introduces a deleterious trade-off by sacrificing source domain performance for unseen targets~\cite{ganin2016domain}, this vulnerability is a critical flaw. It is particularly acute in modern transformers, whose reliance on layer normalization can render them exceptionally sensitive to such distribution shifts~\cite{dosovitskiy2020image}.


To address these limitations, we propose \textbf{Universal Harmonization (U-Harmony)}, a unified module for robust domain-invariant JT in 3D medical image segmentation for DL-based models. U-Harmony introduces a \textbf{two-stage feature harmonization and restoration mechanism} that aligns incoming features with prior distributions and denormalizes them to integrate both new and existing distributions. U-Harmony ensures consistent feature representations across heterogeneous datasets, enabling joint multi-domain training while preserving original dataset-specific variations. Therefore, the priors of original datasets can be explicitly exploited to mitigate the limitations in DG. Additionally, U-Harmony incorporates segmentation models with a \textbf{domain-gated head} that enhances inference for data without indicating the testing dataset, allowing the model to generalize effectively to samples without requiring explicit fine-tuning. 
U-Harmony introduces a superior JT methodology to create a more generalizable and robust segmentation framework, outperforming conventional JT techniques. Our results highlight the effectiveness of U-Harmony in mitigating domain and modality shifts while preserving dataset-specific characteristics.

\begin{figure*}[!t]
\centering
    \includegraphics[width=0.8\linewidth]{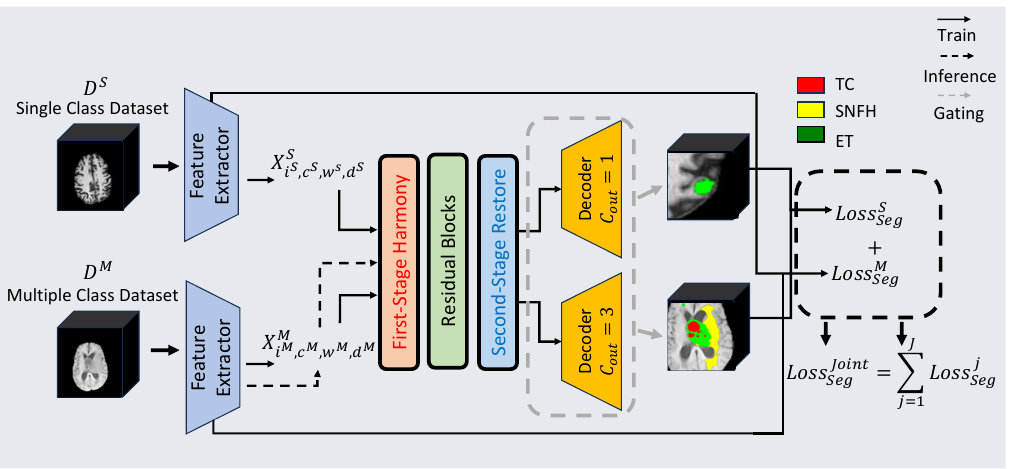} 
    \vspace{-0.4cm}
    \caption{The overview of our U-Harmony framework and domain-gated head network.}
\label{fig:transformer_U-Harmony_workflow}
\end{figure*}

\section{Method}
We propose a universal harmonization module for joint multimodal 3D medical image segmentation across heterogeneous domains, designed to address data shifts from diverse institutions imaging protocols and anatomical structures. The model learns to segment new anatomical structures while adapting to shifts in imaging modalities and acquisition protocols.

Given multiple domains $D^{(j)} = \{(x, s) \mid x \in X^{(j)}, s \in S^{(j)}\}$ with domain-specific inputs $X^{(j)}$ and label sets $S^{(j)}$, we define unified spaces $X = \bigcup_{j \in \mathcal{J}} X^{(j)}$ and $Y = \bigcup_{j \in \mathcal{J}} S^{(j)}$. The model learns $f_{\theta}: X \to Y$ to perform segmentation across all domains. The key challenge is handling \textbf{unaligned class distributions} where $|S^{(j)}| \neq |S^{(k)}|$ and $S^{(j)} \neq S^{(k)}$ across domains, alongside appearance drift in $X^{(j)}$. Our JT framework enables a single model to generalize across diverse datasets with heterogeneous inputs and label compositions without requiring domain-specific models.

\subsection{Joint training with U-Harmony}

Our \textbf{U-Harmony} standardizes feature distributions per instance, reducing sensitivity to dataset variations. Unlike BatchNorm~\cite{wang2022understanding}, which is known to struggle with domain shifts due to its reliance on batch statistics, LayerNorm and InstanceNorm~\cite{instance} operate per-sample and are thus more robust to this issue. U-Harmony ensures domain change by \textbf{harmonizing} instance features in the first stage, utilizing learnable affine transformation, and \textbf{restoring} them in the second stage, enabling knowledge retention without aligning multi-institutional data (Fig.~\ref{fig:transformer_U-Harmony_workflow}).

\noindent \textbf{First-stage Harmonization.}
Given the input feature map of the $i$-th sample \( \bm{x}^{(i)} \in \mathbb{R}^{c \times h \times w \times d} \), we first define the spatial positions $\mathcal{P} = \{p=(h,w,d) \mid 1\leq h\leq H,\,1\leq w\leq W,\,1\leq d\leq D\}$.
In our first stage, the harmonized feature is computed:
\begin{equation}
\hat{\bm{x}}^{(i)}_{c,p}
=
\frac{
\bm{x}^{(i)}_{c,p}
- \displaystyle\frac{1}{|\mathcal{P}|}\sum_{p \in\mathcal{P}} \bm{x}^{(i)}_{c,p}
}{
\sqrt{
\displaystyle\frac{1}{|\mathcal{P}|}\sum_{p\in\mathcal{P}}
\left(\bm{x}_{c,p}^{(i)}
- \displaystyle\frac{1}{|\mathcal{P}|}\sum_{p\in\mathcal{P}} x^{(i)}_{p}\right)^2 + \epsilon
}
}.
\label{eq:comp_norm}
\end{equation}
where \(\epsilon\) is a small constant for numerical stability.

\noindent \textbf{Affine Transformation.}
To compensate for the removed instance-specific information, U-Harmony applies a learnable affine transformation enhanced with higher-order polynomial terms. We get $\hat{\bm{x}}^{(i)}_{c,p}$ is the normalized feature that results from the first-stage harmonization (Eq. \ref{eq:comp_norm}). The variable $j$ is an integer that serves as the exponent for the polynomial terms, ranging from 1 to $J$, where $J$ is a hyperparameter you set that defines the maximum order of the polynomial.  Applying an affine transformation yields $\tilde{\bm{x}}_{c,p}^{(i)}$ as transformed harmonized feature:
\begin{equation}
\tilde{\bm{x}}_{c,p}^{(i)} = w_c \cdot \hat{\bm{x}}^{(i)}_{c,p} + b_c \;+\; \sum_{j=1}^{J} \lambda_{c,j} \cdot \Bigl(\hat{\bm{x}}^{(i)}_{c,p}\Bigr)^j,
\label{eq:comp_affine}
\end{equation}
\vspace{-5pt}

\noindent where \(w_c\) and \(b_c\) are learnable scale and shift parameters, and \(\lambda_{c,j}\) for \(j\in[1,J]\) are additional learnable coefficients that capture higher-order feature interactions.

\noindent \textbf{Second-stage Restoration.}
The key innovation in U-Harmony is the second-stage restoration, which restores critical domain-specific variations. To reintroduce instance-specific variations while preserving the standardized benefits (Eq.~\ref{eq:comp_norm}), we apply a second-stage restoration step using the same statistics $\mu_{i,c}$ and $\sigma_{i,c}$. The restored feature $\hat{\bm{y}}^{(i)}_{c,s}$ is defined as:

\begin{equation}
\hat{\boldsymbol{y}}_{c, p}^{(i)}=\sigma_{i, c}\left(\frac{\tilde{\boldsymbol{x}}_{c, p}^{(i)}-b_c-\sum_{j=1}^J \gamma_{c, j}\left(\tilde{\boldsymbol{x}}_{c, p}^{(i)}\right)^j}{w_c+\epsilon+\sum_{j=1}^J \delta_{c, j}\left(\tilde{\boldsymbol{x}}_{c, p}^{(i)}\right)^j}\right)+\mu_{i, c},
\end{equation}
\label{Eq3}
\vspace{-5pt}

\noindent where $\gamma_{c,j}$ and $\delta_{c,j}$ are additional learnable parameters, and $\epsilon$ is a small constant for numerical stability. This second stage \textbf{selectively} restores domain-specific information $(\mu_{i,c},\,\sigma_{i,c})$ that was removed during normalization while retaining the polynomial transformations previously introduced (Eq.~\ref{eq:comp_affine}). For 3D medical DL-based segmentation models, U-Harmony's first stage harmonization replaces the standard normalization layers to standardize embeddings, and reintroduces this domain-specific knowledge to the decoder with extra second-stage restoration layers. This enables effective learning of shared representations across multi-domain and multimodal datasets while preserving domain-specific information.

\subsection{Domain-Gated Head for Dataset-free Inference}

Rather than training multiple dataset-specific heads, we propose a \textbf{domain-gated head} that jointly learns across $J$ datasets. A backbone $\varphi$ extracts deep features, and a shared head $H$ uses a dataset-adaptive gating function, $G(\varphi(x)) = \mathrm{softmax}(W_G \varphi(x) + B_G)$ (with $W_G \in \mathbb{R}^{J\times m}, B_G \in \mathbb{R}^{J}$), to determine domain contribution. This is guided by domain prototypes $p^{(j)}$, which capture dataset-level feature means via $p^{(j)} = \frac{1}{|X^{(j)}|}\sum_{x\in X^{(j)}} \varphi(x)$. The cosine similarity between the feature and these prototypes, $\mathrm{Sim}(x,p^{(j)}) = \frac{\varphi(x)\cdot p^{(j)}}{|\varphi(x)||p^{(j)}|}$, is then computed to help infer the most relevant domain and dynamically route the features to the correct output path.
At inference, the gating weights and prototype similarities together infer the most relevant domain \textbf{without requiring dataset labels}, enabling consistent segmentation across heterogeneous medical datasets. This inference is achieved by combining the gate's soft domain prediction $G(\varphi(x))$ with the similarity $\mathrm{Sim}(x,p^{(j)})$ to produce a dynamic routing weight. This weight is then used to select or modulate the appropriate decoder path (e.g., `Single Class' vs. `Multiple Class' as shown in Fig.~\ref{fig:transformer_U-Harmony_workflow}), ensuring the model activates the correct output channels for the inferred domain.

\section{Experiments}

\subsection{Experimental setup}
Our method was evaluated on three publicly available brain tumor and metastasis segmentation datasets with varying modalities, complexity, and heterogeneity: (i) UCSF-BMSR (560 MR scans, T1 pre/post-contrast and FLAIR, 5,136 annotated metastases), (ii) BrainMetShare (156 whole-brain MR scans with four modalities: T1 spin-echo pre/post, T1 gradient-echo post, and T2 FLAIR), and (iii) BraTS-METS 2023 (1,303 MR scans with T1 pre/post, T2, and FLAIR, annotated for Enhancing Tumor, Tumor Core, and SNFH). All the datasets were split 70\%/10\%/20\%.  Input volumes were cropped as $96\times96\times 96$ patches and augmented via flipping, rotation, and intensity transformations. Training was performed on an NVIDIA A100 GPU using AdamW (initial LR $1\times10^{-4}$, decay $1\times10^{-5}$) for 50 warmup and 300 total epochs. Code will be released upon acceptance.

\begin{table}[!t]
\centering
\caption{Comparison and ablation studies evaluated with Dice Similarity Coefficient (DSC [\%]). The table shows baseline models and our U-Harmony module applied to nn-UNet and SwinUNETR. Best and second-best results are in \textbf{bold} and \underline{underlined}, respectively.}
\resizebox{\columnwidth}{!}{%
\begin{tabular}{l|c|c|cccc}
\toprule
\multirow{2}{*}{\textbf{Method}} & \textbf{UCSF-BMSR} & \textbf{BrainMetShare} & \multicolumn{4}{c}{\textbf{BraTS-METS 2023}} \\
\cmidrule(lr){2-2} \cmidrule(lr){3-3} \cmidrule(lr){4-7}
& \textbf{ET} & \textbf{ET} & \textbf{TC} & \textbf{SNFH} & \textbf{ET} & \textbf{Average (std)} \\
\midrule
\multicolumn{7}{l}{\textit{Baseline Models}} \\
\quad 3D UNet~\cite{cciccek20163d} & 72.45 & 59.47 & 49.36 & 56.31 & 50.57 & 52.08 $\pm$ 3.03 \\
\quad 3D TransUNet~\cite{chen20233dtrans} & 76.35 & 64.25 & 52.31 & \textbf{67.11} & \textbf{71.43} & \underline{63.61 $\pm$ 8.19} \\
\quad V-Net~\cite{milletari2016vnet} & 73.56 & 62.99 & 56.12 & 59.72 & 60.14 & 58.66 $\pm$ 1.80 \\
\quad UNETR~\cite{hatamizadeh2022unetr} & 72.72 & 60.38 & 62.35 & 57.51 & 56.93 & 58.93 $\pm$ 2.42 \\
\quad nnFormer~\cite{zhou2021nnformer} & 73.98 & 61.73 & 48.34 & 64.17 & 67.73 & 60.08 $\pm$ 8.43 \\
\hline
\multicolumn{7}{l}{\textit{CNN Model}} \\
\quad \textbf{nn-UNet}~\cite{isensee2018nnunet} & 77.41 & 63.96 & 50.25 & \underline{65.92} & \underline{67.86} & 61.34 $\pm$ 7.88 \\
\quad + U-Harmony & \underline{78.15} & \underline{64.41} & 54.63 & 65.71 & 68.59 & 62.98 $\pm$ 5.59 \\
\hline
\multicolumn{7}{l}{\textit{Transformer Model}} \\
\quad \textbf{SwinUNETR}~\cite{hatamizadeh2021swin} & 74.68 & 63.31 & \underline{69.35} & 60.62 & 59.48 & 63.15 $\pm$ 4.41 \\
\quad + U-Harmony & \textbf{80.13} & \textbf{65.63} & \textbf{73.66} & 62.85 & 63.02 & \textbf{66.51 $\pm$ 5.06} \\

\bottomrule
\end{tabular}
}
\label{table:model_performance}
\end{table}

\begin{figure}[!h]
    \includegraphics[width=\columnwidth]{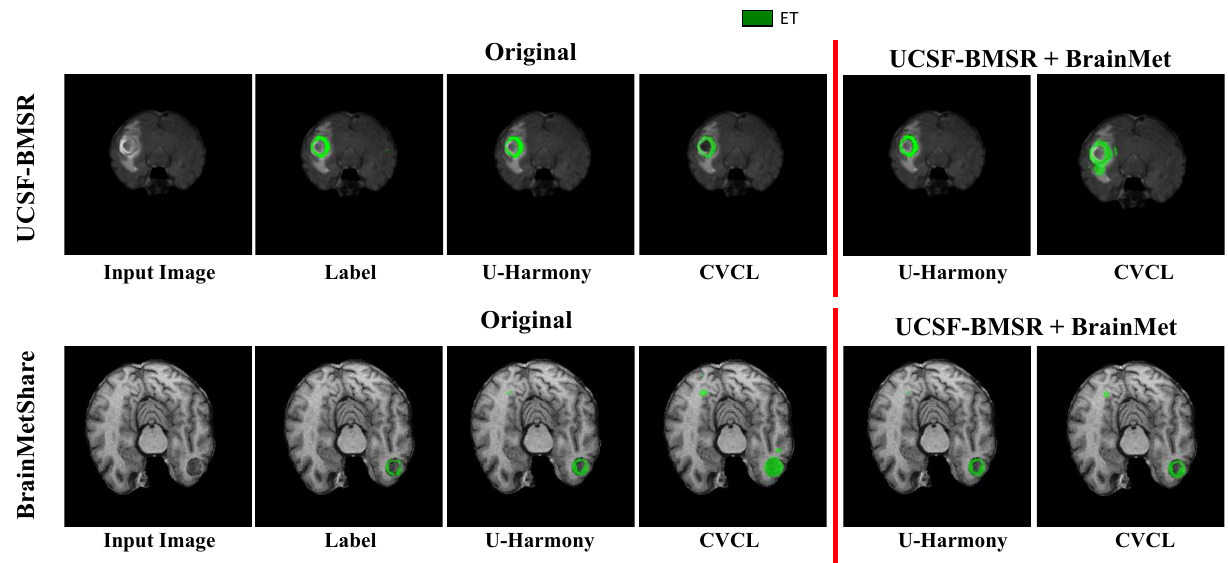}  
    \caption{The comparison showcases the segmentation results on Original datasets UCSF-BMSR and BrainMetShare, transitioning to joint tasks (UCSF-BMSR + BrainMet). The U-Harmony method achieves superior boundary delineation and adaptive accuracy across stages compared to CVCL.}     \label{fig:visual_results3}
\end{figure}

\begin{figure*}[!h]
    \centering
    \includegraphics[width=0.925\linewidth]{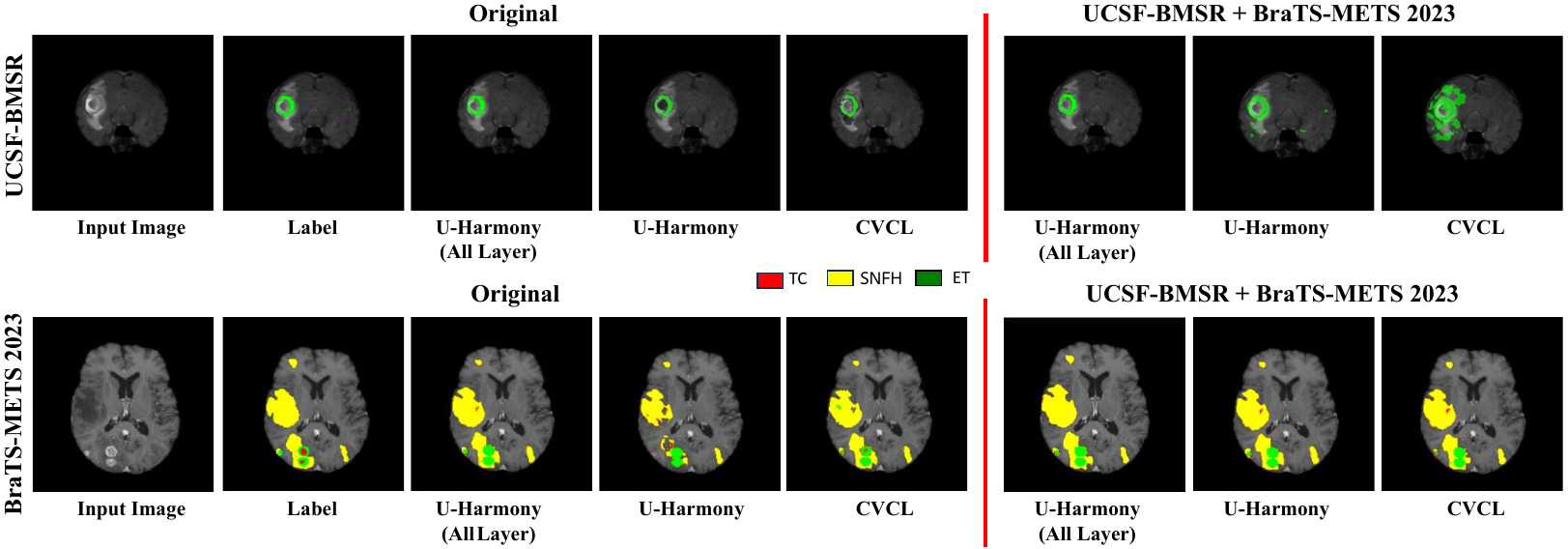}  
    \caption{ The comparison with ablation showcases the segmentation results on Original datasets UCSF-BMSR and BraTS-MENTs 2023, transitioning to joint tasks (UCSF-BMSR + BraTS-MENTs 2023). The highlighted regions demonstrate U-Harmony's robustness in handling domain shifts and extra classes.}     \label{fig:visual_results_v2}
\end{figure*}

\begin{table*}[t!]
    \centering
    \caption{Comparison of Dice Similarity Coefficient (DSC) between different class and domain JT methods with unaligned modalities. Best and second-best performances are marked in \textbf{bold} and \underline{underlined}, respectively.}
    \renewcommand{\arraystretch}{1.2} 
    \setlength{\tabcolsep}{5pt}      
    \resizebox{0.9\textwidth}{!}{%
    \begin{tabular}{@{}l|c|cccc|c|cccc|cc@{}}
        \toprule
        \multirow{3}{*}{\textbf{Method}} & \multicolumn{5}{c|}{\textbf{UCSF-BMSR + BraTS-METS 2023}} & \multicolumn{5}{c|}{\textbf{BrainMet + BraTS-METS 2023}} & \multicolumn{2}{c}{\textbf{UCSF-BMSR + BrainMet}} \\ 
        \cmidrule(lr){2-6} \cmidrule(lr){7-11} \cmidrule(lr){12-13}
         & \textbf{UCSF-BMSR} & \multicolumn{4}{c|}{\textbf{BraTS-METS 2023}} & \textbf{BrainMet} & \multicolumn{4}{c|}{\textbf{BraTS-METS 2023}}   & \textbf{UCSF-BMSR}  & \textbf{BrainMet} \\ 
         \cmidrule(lr){2-2} \cmidrule(lr){3-6} \cmidrule(lr){7-7} \cmidrule(lr){8-11} \cmidrule(lr){12-12} \cmidrule(lr){13-13}
        & \textbf{ET} & \textbf{TC} & \textbf{SNFH} & \textbf{ET} & \textbf{Avg (std)} & \textbf{ET} & \textbf{TC} & \textbf{SNFH} & \textbf{ET} & \textbf{Avg (std)} & \textbf{ET} & \textbf{ET} \\ 
        \midrule
        CVCL~\cite{liu2022context}      & 69.43 & 57.68 & 50.44 & 60.15 & 56.09 $\pm$ 4.12 & 60.39 & 58.02 & 51.27 & 53.97 & 54.42 $\pm$ 2.77 & 76.73 & 61.56 \\ 
        MoME~\cite{zhang2024foundation} & \ulgood{79.12} & \ulgood{64.41} & 60.25 & 67.22 & \ulgood{63.96 $\pm$ 5.61} & \ulgood{67.78} & \underline{63.27} & 62.21 & \underline{67.74} & \ulgood{64.41 $\pm$ 5.12} & \ulgood{79.51} & \ulgood{67.52} \\
        MultiTalent~\cite{ulrich2023multitalent} & 75.35 & 50.44 & \underline{66.15} & 68.02 & 61.54 $\pm$ 5.19 & 64.68 & 48.29 & 63.27 & 66.14 & 58.23 $\pm$ 3.61 & 78.51 & 62.97 \\ 
        \midrule
        nn-UNet                         & 74.18 & 51.33 & 65.12 & \textbf{68.45} & 61.63 $\pm$ 4.59 & 64.15 & 48.37 & \underline{64.22} & 65.59 & 59.39 $\pm$ 4.11 & 74.47 & 64.02 \\
        +U-Harmony                      & 78.31 & 55.68 & \textbf{66.19} & \underline{68.33} & 63.40 $\pm$ 4.11 & 64.91 & 55.02 & \textbf{65.96} & \textbf{68.19} & 63.06 $\pm$ 3.92 & 78.47 & 66.39 \\
        \midrule
        SwinUNETR                       & 69.95 & 57.51 & 62.33 & 61.78 & 60.44 $\pm$ 3.74 & 63.72 & 62.78 & 53.46 & 50.20 & 55.48 $\pm$ 5.17 & 70.55 & 64.23 \\
        +U-Harmony                      & \textbf{83.34}  & \textbf{74.48} & 65.18 & 66.85 & \textbf{68.83 $\pm$ 3.25} &\textbf{70.63} & \textbf{73.79} & 62.37 & 63.67 & \textbf{66.61 $\pm$ 5.13} & \textbf{82.05} & \textbf{71.59} \\
        \bottomrule
    \end{tabular}%
    }
    \label{table:maintable2}
\end{table*}


\subsection{Experimental results}

Our results highlight three key findings:

\noindent{\bf (i) \textit{Robustness to domain shifts.}} 
As shown in Table~\ref{table:model_performance}, U-Harmony outperformed competing segmentation models across all three datasets, achieving the highest Dice scores for enhancing tumor (ET), tumor core (TC), and surrounding non-enhancing FLAIR hyperintensity (SNFH). When integrated with both convolutional (nn-UNet) and transformer-based (SwinUNETR) architectures, U-Harmony provided average DSC improvements of 1.6–3.4\% over the baselines, demonstrating robustness to domain and modality variations.

 \noindent{\bf (ii) \textit{Generalization in cross-domain joint training.}} 
 When trained jointly on datasets with unaligned modalities and heterogeneous annotations, U-Harmony demonstrated stable performance across all source and target domains (Table~\ref{table:maintable2}). Compared to multi-domain baselines such as CVCL, MoME, and MultiTalent, our method achieved higher DSCs (up to +8.4\%) and exhibited superior boundary delineation under severe domain shifts (Figs. \ref{fig:visual_results3} and \ref{fig:visual_results_v2}).
 
\noindent\textbf{(iii) \textit{Effectiveness of U-Harmony and domain-gated head.}} 
To evaluate the effectiveness of our U-Harmony and domain-gated head, we incorporated them into SwinUNETR and compared their performance with other widely used normalization methods and individual components from U-Harmony. Our method consistently outperformed other methods on all the three datasets (see Table~\ref{table:unetr_performance}). Additionally, we found that the more U-Harmony layers are utilized, the better performance we can achieve (see evidence in Fig.~\ref{fig:visual_results_v2}). Finally, to validate our domain-gated, dataset-free design, we compared our full framework against an \textbf{oracle} baseline that uses separate segmentation heads, which requires dataset identification at inference. Although our method still lagged behind the oracle baseline by 1.3\% in DSC (see Table~\ref{table:unetr_performance}), it does not require dataset identification during testing.

\begin{table}[!t]
\caption{Performance comparison of different normalization methods and different components of U-Harmony added to a \textbf{SwinUNETR} baseline, evaluated using DSC. Best and second-best results are in \textbf{bold} and \underline{underlined}.}
\resizebox{\columnwidth}{!}{%
\begin{tabular}{l|c|cccc}
\toprule
\multirow{3}{*}{\textbf{Method}} & \multicolumn{5}{c}{\textbf{UCSF-BMSR + BraTS-METS 2023}} \\ \cmidrule(lr){2-6}
&\textbf{UCSF-BMSR} & \multicolumn{4}{c}{\textbf{BraTS-METS 2023}} \\
\cmidrule(lr){2-2} \cmidrule(lr){3-6} 
 & \textbf{ET}  & \textbf{TC} & \textbf{SNFH} & \textbf{ET} & \textbf{Average (std)} \\
\midrule
+ BN & 63.51  & 59.55 & 51.26 & 57.25 & 56.02 $\pm$ 6.24 \\
+ LN & 74.68  & 69.35 & 60.62 & 59.48 & 63.15 $\pm$ 4.41 \\
+ RevIN~\cite{instance} & 75.13  & 72.45 & 62.01 & 62.58 & 65.68 $\pm$ 2.31 \\
+ U-Harmony (First Stage Only) & 65.41  & 62.17 & 56.52 & 59.25 & 59.31 $\pm$ 5.31 \\
+ U-Harmony (No Affine Transformation) & 76.64  & 63.16 & 57.95 & 62.17 & 61.09 $\pm$ 4.47 \\
+ U-Harmony (2-Layers) & 82.03  & 74.22 & 64.51 & 66.09 & 68.27 $\pm$ 3.95 \\
+ U-Harmony  & \underline{83.34}  & \underline{74.48} & \underline{65.18} & \underline{66.85} & \underline{68.83 $\pm$ 3.25} \\
+ U-Harmony (Multiple Dataset-Specific Heads)  & \textbf{85.31}  & \textbf{75.72} & \textbf{67.52} & \textbf{67.15} & \textbf{70.12 $\pm$ 3.25} \\
\bottomrule
\end{tabular}
} 
\label{table:unetr_performance}
\end{table}
\vspace{-3mm}

\section{Conclusion}
\vspace{-2mm}
We proposed a framework that successfully leverages multiple public medical datasets with diverse appearances and unaligned class structures. Through joint training, this framework trains a unified segmentation model capable of accurate segmentation without requiring dataset identification at inference time. U-Harmony functions as a simple add-on module for both transformer and convolutional backbones (e.g., nnUNet), enabling them to learn generalizable knowledge while preserving unique dataset-specific characteristics. 


\section{COMPLIANCE WITH ETHICAL STANDARDS} 
\vspace{-2mm}
This research study was conducted retrospectively using human subject data made available in open access. Ethical approval was not required as confirmed by the license attached with the open-access data.

\section{Acknowledgement}
This work was partially supported by JSPS KAKENHI, Grant Number JP25K03130.

\bibliographystyle{IEEEbib}
\bibliography{strings,refs}

\end{document}